**Programmable Control of Ultrasound Swarmbots through Reinforcement Learning**

*Matthijs Schrage, Mahmoud Medany, and Daniel Ahmed\**

*M. Scharge, M. Medany, D. Ahmed*
*Acoustic Robotics Systems Lab*
*Institute of Robotics and Intelligent Systems*
*Department of Mechanical and Process Engineering*
*ETH Zurich 8092, Switzerland*
*E-mail: dahmed@ethz.ch*

**Abstract**

Powered by acoustics, existing therapeutic and diagnostic procedures will become less invasive and new methods will become available that have never been available before. Acoustically driven microrobot navigation based on microbubbles is a promising approach for targeted drug delivery. Previous studies have used acoustic techniques to manipulate microbubbles in vitro and in vivo for the delivery of drugs using minimally invasive procedures. Even though many advanced capabilities and sophisticated control have been achieved for acoustically powered microrobots, there remain many challenges that remain to be solved. In order to develop the next generation of intelligent micro/nanorobots, it is highly desirable to conduct accurate identification of the micro-nanorobots and to control their dynamic motion autonomously. Here we use reinforcement learning control strategies to learn the microrobot dynamics and manipulate them through acoustic forces. The result demonstrated for the first time autonomous acoustic navigation of microbubbles in a microfluidic environment. Taking advantage of the benefit of the second radiation force, microbubbles swarm to form a large swarm, which is then driven along the desired trajectory. More than 100 thousand images were used for the training to study the unexpected dynamics of microbubbles. As a result of this work, the microrobots are validated to be controlled, illustrating a good level of robustness and providing computational intelligence to the microrobots, which enables them to navigate independently in an unstructured environment without requiring outside assistance.

**Introduction**

Micro-robotics technology is an integral part of modern-day microsystem technology, and it is being utilized for a variety of biomedical applications ranging from tissue engineering to clinical diagnostics. As a result of its untethered external actuator, the microrobot can reach the majority of locations within the human body and provide accurate diagnoses. Furthermore, the microrobot can provide safe and minimally invasive treatment, therefore reducing the risk of infection and internal injuries.

Precise, accurate, and controlled motion of microrobots can create new opportunities in 3-D manipulation, micro-assembly, sensory applications, cost-effective lithography, and numerous biological and surgical applications. A variety of physical and chemical strategies exist for propulsion at the micro-scale, but their use in bio-applications is limited due to a lack of biocompatibility, low propulsion speed and forces, and poor navigation capabilities. For example, micro/nanorobots powered by chemical fuels[1] and electric[2] fields suffer from poor navigation capabilities and are not entirely biocompatible. Light-induced propulsion is exciting, but typically the propulsive force is low, and implementing in vivo applications with these robots could be a challenge as the light is absorbed or scattered by surrounding tissue [3]Magnetically-powered micro/nanorobots offer the capability for precise navigation[4–11], but many of these robots feature complex 3-D structures that require involved fabrication procedures, and they also have weak propulsive forces. Recently, acoustic field-based propulsion has garnered considerable interest as an alternate means to generate propulsion forces and maneuver micro- and nano-sized objects. Acoustically-induced micro/nanorobots can generate large propulsive forces [12–16] and have good potential for in vivo use; however, they also have poor navigation capabilities and the mechanisms underlying many extant ultrasound-based robots are not well understood.

Among the manipulation methods discussed earlier, acoustic manipulation has emerged as a low-power, biocompatible, and versatile method that overcomes the limitations of other methods. Furthermore, acoustics manipulation techniques for biomedical applications are available within a wide frequency range, ranging from 20 kHz to 16 MHz [17]. Accordingly, acoustic actuation techniques are capable of manipulating agents from clusters of nanoparticles to biological worms measuring up to a few millimeters in length [18]. In addition, this frequency range complements the working range of ultrasound imaging devices, allowing acoustic manipulation techniques to be used in a variety of medical applications. Acoustic manipulation techniques have the advantage of using sound intensities that are less than 10 $W/cm^2$ on the target agents, making them safe for use on biological agents such as embryos and cells [19].

Acoustic microrobots are currently lacking the intelligence required to make independent decisions. This is because the physics underlying microbubble forces is nonlinear, and the known set of governing equations typically require perturbation approaches. It becomes even more difficult to control microrobots when they are submerged in complex liquids such as blood or viscous gel. As a result, path-planning approaches are not viable control methods. In addition, the high speed of the robot and the number of parameters involved in its motion make it extremely challenging for a human controller to accurately predict and manually correct the robot's position in real-time.

In recent years, machine learning has made significant progress in solving stochastic problems and reducing the complexity of physical processes. Many of these strategies have been implemented in robotics and control systems[20–22]. An example of this is reinforcement learning, where the agent is able to learn and gain experience by interacting with its environment. Robots receive rewards or penalties based on each decision they make, whether they move toward the target or away from it. In the entire process, every step from the beginning to the end is referred to as a policy, the robot follows the policy, which is improved through repeated cycles of actions called episodes, resulting in a policy that maximizes cumulative reward. This policy then directs the agent's actions. In recent studies, reinforcement learning has been shown to provide optimal strategies for navigating microswimmers and active particles through fluid flows[23–26], the robot's collective behavior[27–30], and deep learning and deep Q_learning approaches have been used to automate microrobots using magnetic or light sources [31–34].

Micro/nanorobotics platforms based on artificial intelligence are an exciting field of study with a lot of potential for development. While many advanced capabilities and sophisticated control functions have been achieved, the current acoustically powered microrobots still face an unmet challenge in accurately identifying their micro-nanorobots and controlling their dynamic motion in a manner that will lay the foundation for the development of next-generation intelligent micro/nanorobots. Our work aims to utilize microbubbles as wirelessly-controlled micro-actuators in the development of next-generation microrobots that incorporate artificial intelligence.

In this work, we find that using reinforcement learning can indeed manipulate micro-nanorobots with accurate target identification and tolerance definition. This system can transport a single swarm of microrobots to the aimed point defined by users. We have empirically explored and discussed critical factors that affect transport performance with different positioning settings for tolerance values. After training the robot to follow a circle, we attempted to improve its path in every episode. The robot was then programmed to follow the policy that maximized the cumulative reward. ETH was spelled by manipulating the microrobot in order to demonstrate the method.

1. Results and Discussion

As explained in the Introduction, difficulties are often encountered in the field of micro-robotics regarding reproducibility. The scale and simplicity of the experimental setup are central to our approach to ensure experiments are performed successfully and with consistency. The experimental schematic shown in (**Figure 1a**) consists of a microfluidic channel and four piezo transducers (PZTs) that act as actuators. The microfluidic channel acts as an artificial blood vessel made from polydimethylsiloxane (PDMS) in which micro swarms are manipulated. The channel is designed as a square that is aligned with the PZTs, which are glued directly to the sides of the PDMS sample.

A solution containing microbubbles (commercially-available biocompatible gas-filled polymeric-shelled Sonovue bubbles) is introduced into the channel. Without any acoustic actuation, they are still dispersed evenly through the solution, but when subjected to an incident acoustic field, the secondary Bjerknes force brings the bubbles together in swarms[35]. This mechanism is exploited to coalesce the microbubbles at arbitrary positions by means of pseudo-randomly actuating each of the PZTs one at a time using a Mersenne Twister random number generator [36]. An in-house Python pipeline has been developed to facilitate communication between instruments. Iteratively, a model-free control algorithm navigates an arbitrary microswarm through a microfluidic channel from position $p_s$ towards the next target point $p_t \in P_t$ along an arbitrary path $P_t \subseteq P_s \in \mathbb{N}^{300 \times 300}$. The goal condition

$$G[n] = \begin{cases} True \leftrightarrow ||p_t - p_s||^2 \leq \delta \; \forall p_t \in P_t \\ False \leftrightarrow else \end{cases} \quad (1)$$

is satisfied when the swarm has consecutively passed all target points that make up the target path. In this notation, $\delta$ denotes the minimum distance that a swarm has to be from one target point in $P_t$ in order to be allowed to move to the next.

The control algorithm predicts optimal values for the output variables consisting of peak-to-peak voltage $V_{pp}$, frequency $f$, and PZT index $k$. That is for every step $n$, the algorithm has to choose an optimal $V_{pp}$ and $f$ for each $k$. Each step is finalized by sending an AC voltage at $V_{pp} \in [0, 20] \; V$ and frequency $f \in [0, 5] \; MHz$ to each of the four PZTs $k \in \{1, 2, 3, 4\}$. The three output variables thus form a 3-dimensional action space, the boundaries of which are set by the limits of the Tektronix AFG3011C function generator and the number of PZTs. Given that this action space is in fact very large, we performed action space reduction in order to speed up the computation time and reduce complexity. A summary overview of the pipeline is illustrated in (**Figure 1c**).

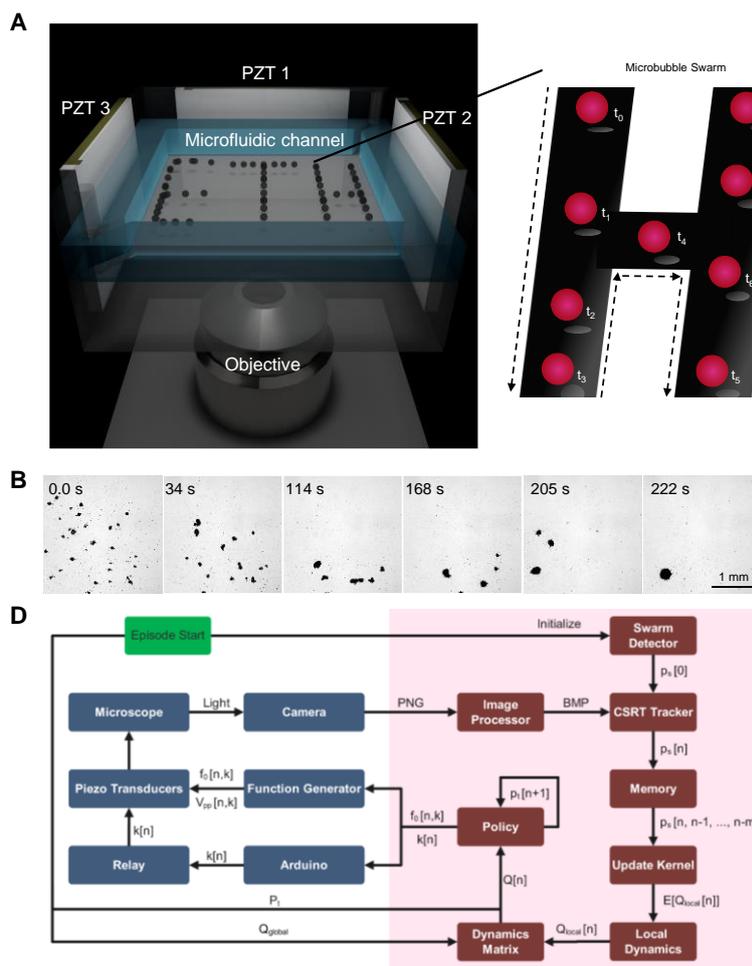

**Figure 1. A)** aschematic for the microscopic Mirorobots setup. A $400 * 400 \mu m$ square channel and $50 \; \mu m$. Each of the walls is attached with a piezo transducers which allow the robot to move in all directions. **B)** From left to right, freshly injected microbubbles are autonomously guided to form a swarm. **C)** Sequential flow diagram of algorithm (on the right in dark red) and instruments (on the left in blue). A swarm detection algorithm, explained in the 'Object detection and tracking' section, initializes a CSRT tracker, which tracks the swarm at location $\boldsymbol{p_s}[n]$ for any $n > 0$. A list of past locations $[\boldsymbol{p_s}[n], \boldsymbol{p_s}[n-1], ..., \boldsymbol{p_s}[n-m]]$ is continuously held in memory to update the learning dynamics matrix $\boldsymbol{Q_{local}}[n]$. This is combined with the static dynamics matrix $\boldsymbol{Q_{global}}$ (loaded from memory at the start of the experiment together with target path $P_t$) to form the combined dynamics matrix $\boldsymbol{Q}[n]$. Policy $\pi$ uses $\boldsymbol{Q}[n]$ to choose the optimal set of outputs based on current position $\boldsymbol{p_s}[n]$ and target position $\boldsymbol{p_t}[n]$. These inputs are transferred to the PZTs, which produce ultrasound at voltage $V_{pp}[n,k]$ and resonant frequency $f_0[n,k]$ to each of the PZTs. This produces a response from the microswarms, which is observed by an observation module through a microscope, after which the process repeats for step $n+1$.

### 1.1. Dynamics of an acoustic micro swarm

As part of the experimental setup, a typical transducer is brought into contact with the PDMS structure to be excited through the use of a coupling agent or glue. The ultrasound waves travel through the PDMS and reach the square environment that contains the microbubbles. As the microbubbles are exposed to the sound field, they oscillate and scatter, giving rise to the second Bjerknes force. Accordingly, the adjacent micro bubbles cluster into a swarm, as depicted in (**Figure 1B**). There appear to be two distinct classes of Bjerknes forces. First, there is the radiation force (first Bjerknes), and second, there is the mutual attraction force (second Bjerknes). As should be understood here, there is only one force present, which is the radiation force that is generated as a result of the acoustic field pressure gradient. Through the action of the primary force, the individual pulsating bubbles generate a secondary sound field which attracts the bubbles together [37]. Detailed explanations can be found in the supporting information.

### 2.2. Action space

We prune the action space from three dimensions to one dimension to simplify the navigation problem and increase the algorithm's precision and convergence speed. There are three reasons that this pruning step is necessary to achieve our results. Firstly, during initial experiments we found the action space that results from the natural limits in our equipment, specifically the Tektronix AFG3011C function generator, to be too large for the learning control algorithm to converge in an acceptable finite timespan without any other explicit constraints. Thus, algorithms that search the three-dimensional action space for an optimal action had difficulty convergent into consistent navigational behavior. Secondly, not only are the natural boundaries too expansive, but a large part of the range of $f$ is completely irrelevant in practice. For one, the acoustic field strength that we require can only be created by PZTs when using frequencies at or around certain resonance frequencies [38] one of which is the optimal theoretical resonance frequency $f_0$ of $\sim 2\ MHz$ for the PZTs used in our study. Thirdly, we experimentally found that small changes in $f$ can produce large differences in the swarm-PZT dynamics of the system; hence, by constraining $f$ we increase reproducibility.

We devised a novel approach to achieve consistent navigation, which is largely enabled by reducing the action space size through pruning. We approached this pruning with the assumption that the microswarm should move at maximum speed at all times, because our algorithm aims to satisfy the goal condition as fast as possible. Thus, we can apply limitations to $V_{pp}$ and $f$ that promote convergence towards the goal condition. In particular, we can experimentally determine the optimal $V_{pp}$ and $f$ for each PZT $k$. This reduces the action space to a one-dimensional $k$ and its corresponding pre-determined $V_{pp}[k]$ and $f[k]$.

Accordingly, we conducted a grid-search of the action space, collected a series of images, and measured the movement of swarms at arbitrary positions in the channel. To create the search set, we first discretized $V_{pp}[n, k] \in [10, 20]$ to form one axis consisting of six values with step size of 2: $V_{pp}[n, k] \in \{10, 12, 14, 16, 18, 20\}$. We then discretized $f[n, k] \in [1.5, 2.5]$, creating a set of 41 values with steps of size 0.025. These were combined with $k[n] \in \{1, 2, 3, 4\}$ to form a search set having $6 \times 41 \times 4 = 984$ samples. The swarm dynamics at each point in the search set was recorded for three seconds at 33 frames per second, producing a dataset of $984 * 3 * 33 = 97{,}416$ images. We collected the recordings over the course of 41 experiments at distinct frequencies, as swarms become less responsive during the course of one experiment. This produced high-quality and reproducible data from which we could make assumptions about swarm-PZT dynamics. Sample footage of these recordings is provided as (**Video S1**) in the Supplementary Information.

Our approach provides a means of rigorously measuring the dynamics of an acoustic microswarm within an acoustic microfluidic system without any prior assumptions. Importantly, the lack of prior assumptions about the swarm dynamics means it could be scalable to *in vivo* applications. Moreover, our results clearly indicate the resonance frequency $f_0[k]$ for each of the four PZTs in the system (**Fig. 3B-E**) as well as an approximately linear relationship between voltage and swarm velocity (**Fig. 4**). Using that data, we applied the following constraints to the action space based the assumption that maximizing the velocity of a swarm allows it to satisfy the goal condition as fast as possible.

$$
\begin{aligned}
&V_{pp}[n, k] \in \{0, 20\}, \forall n, k \\
&\sum_{k=1}^{4} V_{pp}[n, k] = 20 \ \forall n, k \\
&f[n, k] = f_0[k] \ \forall n, k
\end{aligned}
\tag{2}
$$

In other words, policy $\pi$ only has to find a single PZT with index $k = k_n$ to activate at its resonance frequency $f[n, k = k_n] = f_0[k_n]$ and maximum voltage $V_{pp}[n, k = k_n] = 20$ while providing the other three PZTs with $f[n, k \neq k_n] = 0$ and $V_{pp}[n, k \neq k_n] = 0$. By making the input variables dependent or constant, we simplify the control problem to a choice between only four possible actions. This allows us to achieve reproducible navigation of microswarms.

**2.3 Reinforcement learning implementaion**

Our control algorithm is based on a subcategory of reinforcement learning termed 'Q-learning,' originally devised by Watkins [39], which aims to learn the value of an action in any arbitrary state. It is designed to find the optimal policy $\pi^*$ that maximizes the expected value of the total future reward $\mathbb{E}(R)$ for any finite Markov decision process [40]. In short, the agent attempts an action under a given state, for which it receives a reward or penalty. It then assesses that consequence and makes a new attempt. Through repeatedly trying all actions under all conditions, the agent ultimately learns which perform best according to the overall reward.

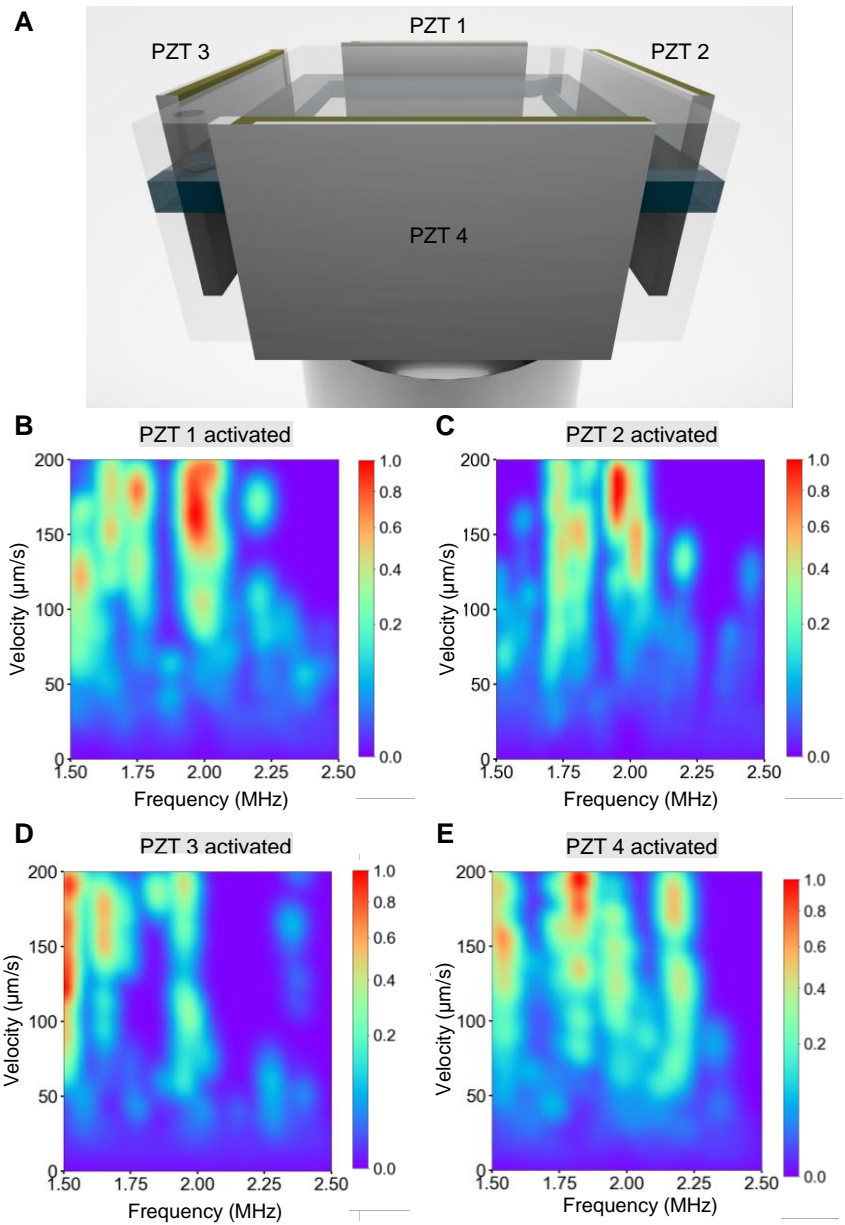

**Figure 2. A)** Schematic of the microfluidic device with four Piezo transducers glued to each surface of the PDMS channel. **B-E)** respectively represent the velocity responses from PZTs 1-4, and were used to determine each PZT's resonance frequency, which corresponds to the frequency having the highest velocity-weighted sample density. Of particular note is in **D)**, which shows a clear shift in resonance frequency relative to both the other PZTs and the manufacturer's specification of $2\ MHz$. This is in line with our experimental observations. We expect the deviation was caused by an error in attaching the PZT to the PDMS, such as extra amount of glue.

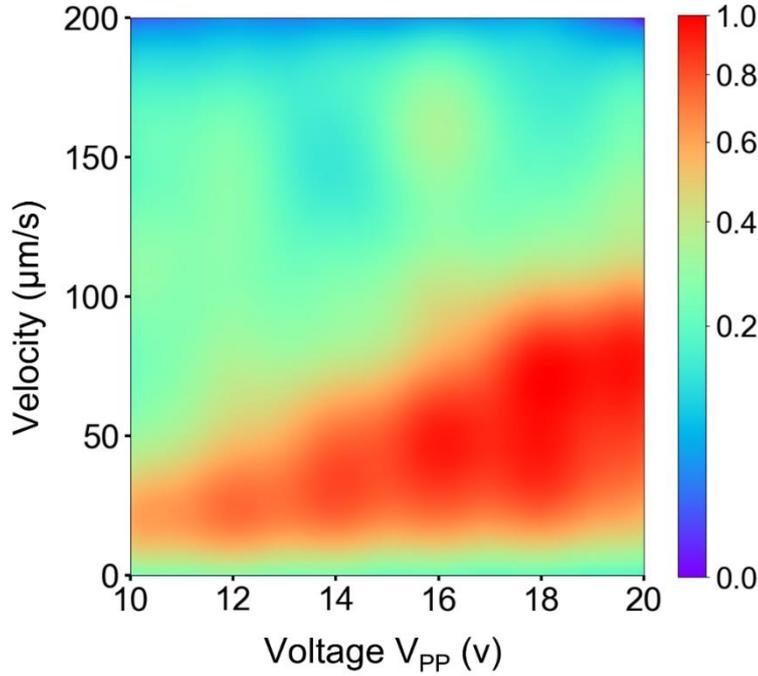

**Figure 3**. Velocity response results obtained during a grid-search of the action space. shows the velocity response to input voltage. Higher peak-to-peak voltage corresponds to a higher velocity, which confirms experimental observations. The search set is shaped as a 4×6×41 array and the swarm responce from each point in the set was measured for three seconds at 33 frames per second. Colors represent sample density, with red areas having the highest values. Samples were evenly distributed on all three variables of the action space ($k, f, V_{pp}$) and swarm velocity was sampled to create a uniform distribution.

To understand the reason for our approach, it must first be understood that the behavior of the microswarm-PZT system is extremely spatially non-linear[41]. Even in our simple experimental setup, the microswarm does not operate in a static environment. In the course of our experiments, auxiliary microbubbles and swarms are present in the channel with spatial and temporal variability, as are contaminants such as dust particles and an increasing number of burst microbubbles as the experiment proceeds. These result in a changing non-linear bulk acoustic wavefront combined with secondary and higher order waves that are time-dependent and highly sensitive to the initial state. This spatially and temporally non-linear behavior makes it difficult for navigation based on a continuously learning algorithm, because convergence is not guaranteed. Such dynamism might also be one of the reasons why theoretical models and simulated models often are not validated by actual acoustic manipulation experiments.

### 2.3.1. Global dynamics

As outlined above, control and automation of acoustic robotic systems has remained a fundamental challenge[42], mainly one of reliability and scalability as *in vivo* applications move towards more non-linear environments. To start to overcome this fundamental limitation, we created a global dynamics matrix $Q_{global}$ for each of the four PZTs; this matrix represents the expected movement $\mathbb{E}[\dot{p}_s[n+1]]$ in $\mu m s^{-1}$ of swarms at all locations in the channel, visualized in **Fig. 4A-D**. The dynamics of the swarms can be extracted from the dataset described earlier, using linear regression to extrapolate a uniform grid of $\mathbb{E}[\dot{p}_s[n+1]] \; \forall p_s \in P_s$ from the non-uniform data.

We specifically extrapolated the dataset to a four-dimensional array, in which two dimensions represent the position $p_s$ in the channel, the third dimension stores $\mathbb{E}[\dot{p}_s[n+1]]$, and the last dimension distinguishes the four PZTs. $Q_{global}$ was then used to perform navigation of microswarms along an arbitrary path. For demonstration purposes, we had the swarms write 'ARSL' as depicted in (**Figure. 6A-D**). Video footage of these experiments is available in (**Video S2**) in the Supplementary Information.

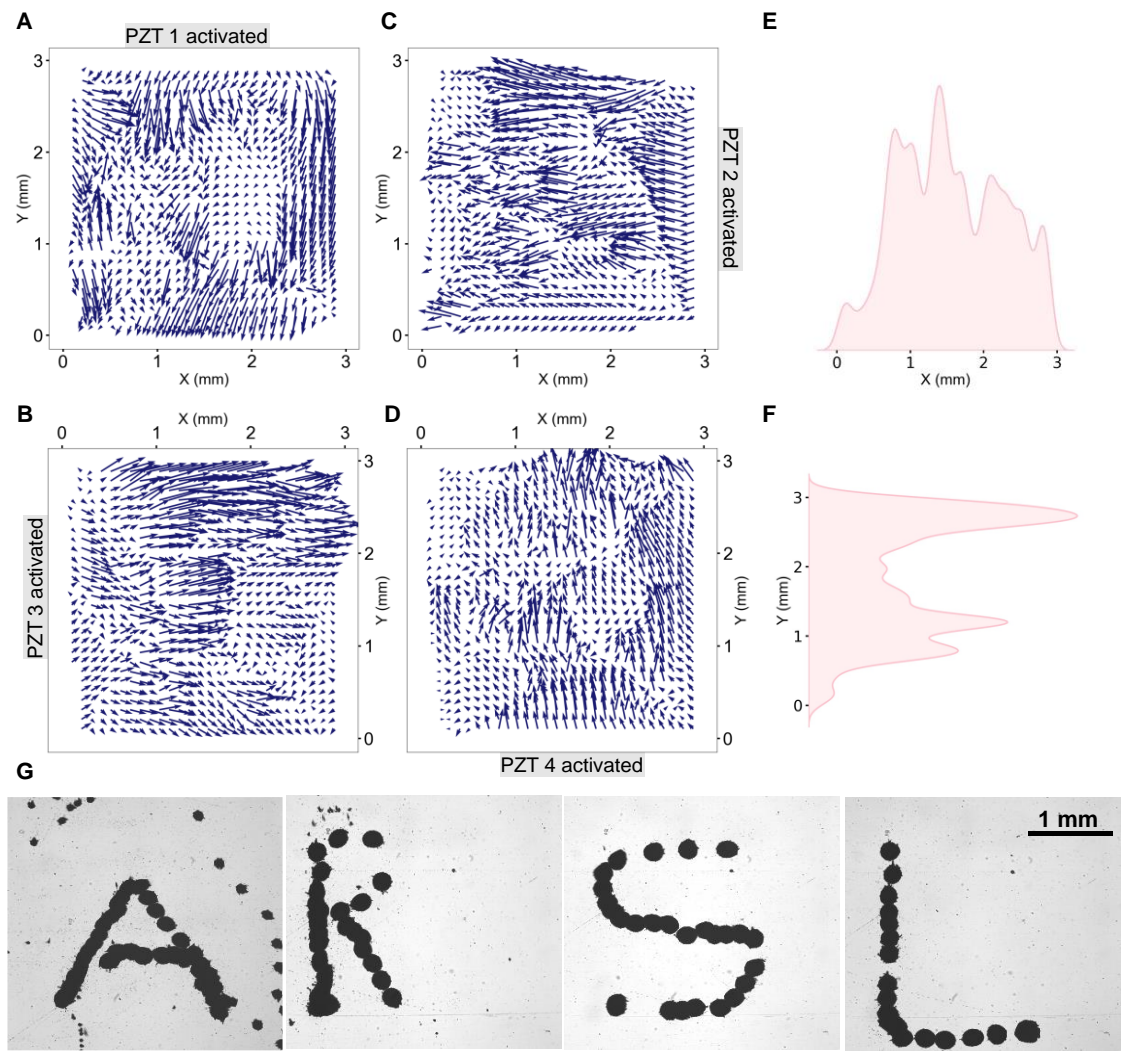

**Figure 4**. **A-D)** The respective global dynamics of PZTs 1-4 were averaged over 100k images. Rows show the predicted movement of an arbitrary microswarm during one second as a function of the swarm's position in the channel. As most arrows point away from the PZTs, it is very evident that the primary bulk acoustic wave is dominant over secondary and higher-order waves. Irregularities in the field can mostly be attributed to the fact that the experimental data is not perfectly uniform, especially at channel edges. In addition, the primary wave is also not expected to be completely uniform because of irregularities in the environment such as dust particles and auxiliary microbubbles and swarms. In both **E)** and **F)** we show the average

distribution of microbubbles injected in the microfluidic channel over all the data in the x and y directions respectivly. **G)** Swarm manipulation using only the global dynamics matrix. The algorithm is given a path consisting of a series of points in a 300-by-300 discretized coordinate system within the channel. It iteratively tries to navigate the centroid of the swarm to consecutive points on the path. It moves on to the next point once the location of the swarm has satisfied the condition $||\boldsymbol{p}_t[n] - \boldsymbol{p}_s[n]||^2 \leq 5 \ \forall \boldsymbol{p}_t$.

### 2.3.2. Local dynamics

We found experimentally that when using only global dynamics, swarms often get stuck in local minima. This is likely the result of the overgeneralization that the global dynamics represent; an algorithm that relies upon generalized behavior in a spatially and temporally dynamic environment is likely to get stuck. Thus, while global dynamics are necessary for navigation, they need to be supplemented by a local dynamics matrix $\boldsymbol{Q}_{local}$ to account for changes in the environment. We used a uniformly initialized matrix of the same shape as $\boldsymbol{Q}_{global}$ for $n = 0$ and use the following formula to update $\boldsymbol{Q}_{local}$ after initialization.

$$\boldsymbol{Q}_{local}[n+1] = (1-\alpha) * \boldsymbol{Q}_{local}[n] + \mathbb{E}[\alpha * \boldsymbol{Q}_{local}[n+1]] \tag{9}$$

where $\alpha \in [0,1]$, the learning rate, can be used to make the function more or less sensitive to changes in dynamics. $\mathbb{E}[\alpha * \boldsymbol{Q}_{local}[n+1]]$ is calculated from the average of a past number of steps collected in memory. To demonstrate the capability of the learning algorithm, we used $\boldsymbol{Q}_{local}$ to direct swarms along an arbitrary path (three laps of a circle) and compared its prediction error to the use of $\boldsymbol{Q}_{global}$ alone (**Figure. 5A-B**). **Video S3** Supplementary Information shows a live visualization of the four-dimensional $\boldsymbol{Q}_{local}$ on the left, and a microswarm navigating the circular path on the right. The experiment used $\alpha = 0.05$.

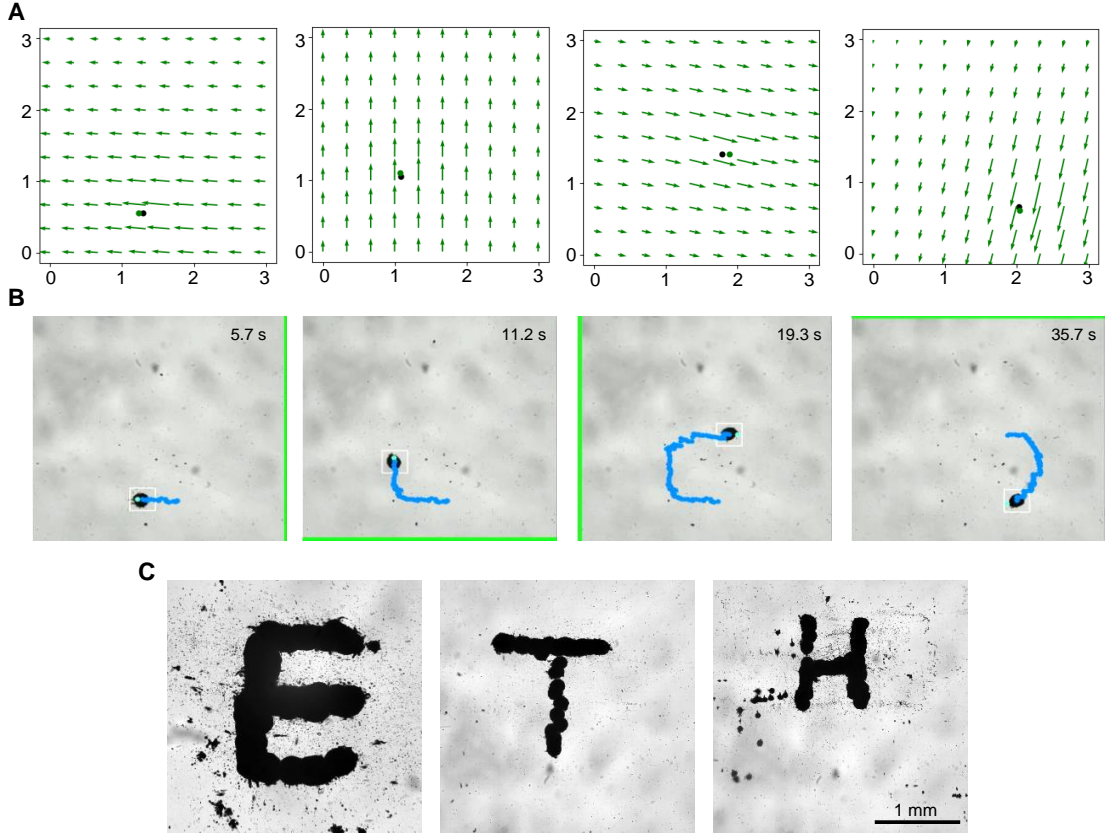

**Figure 5.)** For the purpose of demonstrating the capabilities of the learning algorithm, we used the following algorithm $Q_{local}$ to direct swarms along an arbitrary path. **A)** displays a representation of the swarm moving in a circle in response to the piezo transducer that is being activated. **B)** displays the output of the experiment with real-time manipulation of the swarm to move in a circle. **C)** Swarm manipulation using both global and local dynamics to spill ETH. The demonstration shows the ability of this combined approach to achieve precise manipulation. These experiments had a much higher success rate than those using only global dynamics.

### 2.3.3. Policy

$Q_{global}$ and $Q_{local}$ are combined into one matrix encapsulating overall dynamics using the following formula:

$$Q[n] = \beta * Q_{global} + (1-\beta) * Q_{local}[n] \tag{10}$$

in which $\beta \in [0,1]$ represents the bias towards global behavior. A large $\beta$ makes navigation easier, but increases the chance of the swarm getting stuck in a local minimum. The obtained matrix $Q[n]$ contains the final expected velocity $\mathbb{E}[\dot{p}_s[n+1]]$ for any position and any $k$. When navigating a swarm, the policy finds a $k_{optimal}$ such that

$$k_{optimal}[n=m] = {_{k=1}^{4}}\text{argmax}\,(||p_t[x=a, y=b, n=m] \\ -p_s[x=c, y=d, n=m] - Q[x=c, y=d, n=m+1]||^2 \tag{11}$$

In other words, the policy chooses to activate the PZT with index $k$ at arbitrary step $n = m$ to direct the microswarm at position $x = c, y = d$ if and only if doing so minimizes the expected distance between the swarm and the target position at $x = a, y = b$. To demonstrate the method is capable of precise navigation along an arbitrary path, we used swarms to spell out 'ETH' (**Figure 5C; Video S4 supporting inforamtion**). We found that microswarms manipulated by this policy are less prone to becoming stuck in local minima while still being able to navigate long distances efficiently.

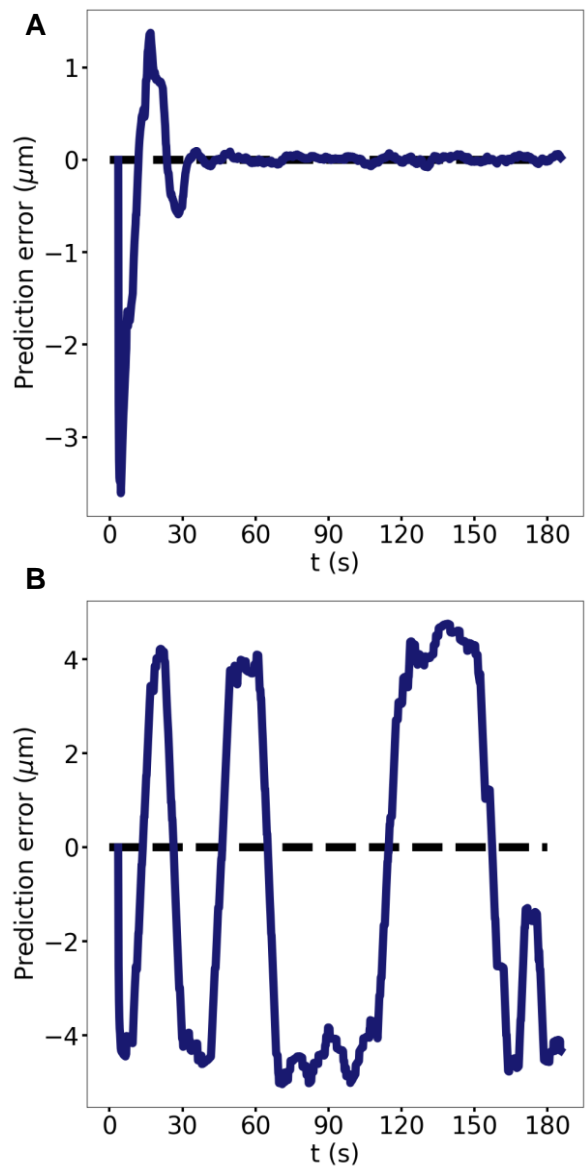

**Figure 6. A)** Local prediction error of a swarm following an arbitrary circular path. The error is defined as the average error of the two-dimensional $\mathbb{E}[\dot{p}_s[n+1]]$ and thus contains information about both angular prediction error and velocity prediction error. The graph starts similarly to an exponentially decaying harmonic wave. **B)** Global prediction error of a swarm following the same arbitrary circular path. The error magnitude does not go down over the course of the experiment.

## 2. Conclusion

In summary, we developed a manipulation platform fully controlled over Python from ultrasound forces actuation to closed-loop feedback. This is done by grabbing and analyzing the images and then detecting and tracking the microrobots. We then used reinforcement learning to enable continuous control of the microrobots and understand their dynamics. Having collected a large amount of experimental data for training, the microrobots were able to move efficiently and autonomously perform tasks and as a demonstration, we used the microswarm to spell ARSL and ETH. Due to the safety and biocompatibility of ultrasound and microbubbles, this work has a high potential for in vivo application.

## 3. Experimental section

The experimental setup is shown in the (**Figure S1 supporting information**) which includes a function generator, observation module, inverted microscope, microcontroller, electronic circuit, microfluidic channel, and piezo transducers (PZTs) that serve as actuators, which were glued to the PDMS wall with two-component epoxy glue. The ultrasound piezo transducer used has a resonance frequency of 2 MHz. A function generator (Tektronix AFG31000), a microcontroller (Arduino-Uno), and a relay module, and a camera (Hamamatsu C11440) connected to an inverted microscope (Leica DMI6000B) were connected to In-house Python code. Images are captured at 33 frames per second by the camera and are processed by Python for autonomous manipulation. Additionally, the Python code controls the frequency, amplitude, and state of the piezo transducers.

Microfluidic channels were fabricated with polydimethylsiloxane (PDMS) using standard soft lithography. A master mold was used to fabricate each device, which was then lithographically patterned with SU-8 negative photoresist on a 4-inch silicon wafer and then placed inside a Petri dish. PDMS prepolymer was prepared by mixing the silicon elastomer base and curing agent at a 10:1 weight ratio, following which the PDMS prepolymers were degassed under vacuum and cast into the mold. PDMS was cured by heat treatment at 85 °C for two hours. The cured PDMS was cut and peeled off from the channel mold, following which the inlet and outlet access ports were created by using a 1-mm-diameter punch. Afterwards, a blank wafer is used to produce a flat layer of PDMS alongside the previously mentioned layer using the same process. The layers are cleaned with isopropyl alcohol and submerged in an ultrasonic water bath for 15 minutes. Next, the PDMS channel was bonded with another layer of PDMS under 85 °C for two hours. Bonding was conducted after plasma pretreatment for 1 min.

Microbubble Contrast Agents: The microbubbles used for manipulation were purchased from Bracco Sonovue imaging contrast agents. The Sonovue contrast agent fabrication kit consists of a glass vial containing (25 mg) lyophilised sulphur hexafluoride lipid-type A powder, one syringe containing (5 ml) sodium chloride (saline) solution. To test different microbubble concentrations, different dilutions of the supplied saline solution were made. The microbubble solution was injected using a pipette into the microfluidic channel. The self-assembly of the microswarms under acoustic forces was recorded by a camera connected to an inverted microscope.

Images analysis of the experiments were processed using an autonomous object detection and tracking system. Our experiments aim to move acoustic microswarm manipulation closer to in-vivo applications which require a robust system to subtract the background from images and analyze them in real time. Automatic detection and tracking systems are capable of processing a lot more information than a manual operator, even if they are very skilled. An operator may be unable to perform acoustic microfluidic experiments with rapid changes in behavior. We therefore developed a feedback system that detects and tracks the location of any swarm in-frame of at least $10\ \mu m$ in diameter at a speed of approximately $50\ FPS$. The minimum swarm diameter is only limited by the microscope magnification and resolution, which in turn is limited by the required processing speed. The visual sensory system consists of a microscope and an observation module and that collects a 2048-by-2048 16-bit .PNG image, $I_{PNG}$, through a 5x magnifying lens to a processing pipeline in the Python programming language in PyCharm 2020.3.3. The pipeline compresses the image to 300-by-300 8-bit grayscale .BMP image, $I_{BMP}$ shown in (**Figure S3A**,Supporting information). This compression decreases timestep $t[n] - t[n+1]$ considerably compared to using a non-compressed image. Not only can the swarm location $\boldsymbol{p}_s$ be extracted quicker, but the size of state space $P_s$ is considerably decreased, which increases convergence speed of our algorithm. For $n = 0$, a threshold condition is applied to $I_{BMP}$ to form a binary image, $I_{thresh}$ as shown in (**Figure S3B**,Supporting information). The threshold boundaries are determined from the distribution (**Figure S3E**,Supporting information). of the pixel intensity of a sample set of images. The intensity threshold filters out any pixels of higher intensity than a specific value and provides maximum contrast between microbubbles, black, and the bulk surroundings, white, but does not yet distinguish between swarms and contaminants. The binary image is then convoluted with a $2 \times 2$ Gaussian blur kernel as shown in (**Figure S3C**,Supporting information). in order to improve accuracy in the next step. Standard Canny edge detection [43] from the OpenCV image processing library was used to produce $I_{canny}$, an image containing the edges of $I_{thresh}$. $I_{canny}$ retains only edge features as shown in (**Figure S3D**,Supporting information).which are used to find the contours of an image using the OpenCV implementation of topological structural analysis [44]. We experimentally found that this implementation inherently filters out almost all non-microbubble contaminants such as dust particles due to their small scale ($\approx 10 - 50\ \mu m$) compared to the microswarms we use ($50 - 200\ \mu m$). On top of this, the OpenCV implementation of topological structural analysis only detects shapes with relatively soft curves, a property that is often not associated with dust particles. If no manual specification is presented to the algorithm, it automatically chooses the one with the largest moment to find $\boldsymbol{p}_s[0]$. For any sequential steps $n > 0$, tracking is used instead of detection to improve processing speed and ensure a continuity of $\boldsymbol{p}_s$. We developed the tracker based on a dataset of recorded footage of microswarms in a similar experiment [45]. The one that we experimentally found to have the best combination of accuracy and speed was the OpenCV implementation of a discriminative correlation filter with channel and spatial reliability [36].

**Funding**. This project has received funding from the European Research Council (ERC) under the European Union's Horizon 2020 research and innovation programme grant agreement No 853309 (SONOBOTS) and ETH Research Grant ETH-08 20-1. We thank Alexia Del Campo Fonseca Dr. S. Karthik Mukkavilli for helpful discussion. **Competing Interest Statement**. The authors declare no competing financial interests. Data and materials availability. All data are available in the main text or the supplementary materials.